\documentclass[a4paper, final]{article}   

\usepackage[margin=1.5in]{geometry}
\usepackage[latin1]{inputenc}  
\usepackage[T1]{fontenc}
\usepackage[ngerman,english]{babel}
\usepackage[english=american]{csquotes}

\usepackage{graphicx}

\usepackage{amsmath}
\usepackage{amsthm}
\usepackage{amssymb}
\usepackage{cleveref}

\usepackage{authblk}  
\title{A robustness measure for singular point and index estimation in discretized orientation and vector fields}

\author[1,2,3,$\ast$]{Karl B.~Hoffmann}
\author[1,2,3,$\ast$]{Ivo F.~Sbalzarini}
\affil[1]{Technische Universit{\"a}t Dresden, Faculty of Computer Science, Dresden, Germany}
\affil[2]{Max Planck Institute of Molecular Cell Biology and Genetics, Dresden, Germany}
\affil[3]{Center for Systems Biology Dresden, Dresden, Germany \newline
} 
\affil[$\ast$]{Email: {karlhoff@mpi-cbg.de}, {ivos@mpi-cbg.de} \newline 
Website: {http://mosaic.mpi-cbg.de} \newline
}

\date{ \today }

\newcommand{\markdef}[1]{\emph{#1}}  

\newcommand{\bb}[1]{\mathbf{#1}}    

\usepackage{amssymb}
\newcommand{\R}{\mathbb{R}}
\newcommand{\Z}{\mathbb{Z}}

\DeclareMathOperator{\modpi}{mod_{\pi}}   

\newcommand{\define}{\mathrel{\mathop:}=} 

\begin{document}

\maketitle

\begin{abstract}

The identification of singular points or topological defects in discretized vector fields occurs in diverse areas ranging from the polarization of the cosmic microwave background to liquid crystals to fingerprint recognition and bio-medical imaging. 
Due to their discrete nature, defects and their topological charge cannot depend continuously on each single vector, but they discontinuously change as soon as a vector changes by more than a threshold. 
Considering this threshold of admissible change at the level of vectors, we develop a robustness measure for discrete defect estimators. 
Here, we compare different template paths for defect estimation in discretized vector or orientation fields. 
Sampling prototypical vector field patterns around defects shows that the robustness increases with the length of template path, but less so in the presence of noise on the vectors. 
We therefore find an optimal trade-off between resolution and robustness against noise for relatively small templates, except for the ``single pixel'' defect analysis, which cannot exclude zero robustness.
The presented robustness measure paves the way for uncertainty quantification of defects in discretized vector fields.


\end{abstract}

\newpage
\section{Introduction: singular points and their robustness}
A \markdef{topological defect} or \markdef{singular point} in a two-dimensional vector field is an isolated discontinuity of an otherwise continuous vector field. 
Such objects can be defined in both vector fields and orientation (i.e., ``headless vector'') fields. 
Without loss of generality, we here only consider the more general case of orientation fields $X \to \R^2_{\sim} $,  
where $ \R^2_{\sim} $ is the set of nematic vectors obtained from $\R^2$ by identifying antipodal polar vectors $\bb{x} , -\bb{x} \in \R^2$ to one nematic vector~$\in \R^2_{\sim}$. 
Topological defects in discretized orientation fields are identified together with their \markdef{topological charge} or \markdef{index} from the cumulated orientation change along closed paths around the tentative defect~\cite{BazenGerez2002,BlowThampiYeomans2014,DoostmohammadiThampiYeomans2016}.

As the topological charge of a defect in an orientation field can only have discrete values~$\in \left\{0, \pm 1/2, \pm 1, \pm 3/2, \mathellipsis \right\}$, it \emph{cannot} depend on each of the input orientations in a globally continuous manner. 
Hence, the charge, location, and existence of a defect can discontinuously change under arbitrarily small perturbations to the orientation field, for example due to temporal dynamics or measurement errors.
To compensate for this, defect identification in discretized orientation fields often uses closed paths larger than the smallest possible loop around a single pixel. 
This increases robustness to noise at the expense of localization resolution~\cite{BazenGerez2002,DeCampRednerBaskaranHaganDogic2015,HutererVachaspati2005}.

\section{A measure of singular point robustness}
To capture the role of noise in defect estimation, 
and to quantify the trade-off between robustness and resolution,
we propose a \markdef{robustness measure} for singular point estimators. 
Our measure quantifies the maximum orientation change admissible at the level of individual vectors \emph{without} changing the topological charge estimated around a given closed path. 
In discretized orientation fields, each edge of the path connects two neighboring discretization points ($\bb{x}_i, \bb{x}_j$) and contributes a finite orientation change $\Delta \theta = \theta_j - \theta_i$ between the orientation angles $\theta_i, \theta_j$ at the two points.
Since orientation angles are $\pi$-periodic ($2\pi$ for vectors), 
the difference $\Delta \theta$ is unique up to additive multiples of $\pi$ (or $2\pi$). 
Thereby, one has to choose the representative $\modpi \left( \Delta \theta \right) \define  \Delta \theta -  \pi \lfloor 0.5 + \Delta \theta/\pi \rfloor \in \left[ \pi/2, \pi/2 \right)$ for consistency with the continuous case in the limit of the discretization grid spacing $h \to 0$. 
Due to the discontinuities of $\modpi$,
the defect estimate changes whenever any orientation angle difference $\theta_j - \theta_i$ crosses~$\pi/2 + k \pi$, $k \in \Z$.
We define the robustness of a defect estimator as the largest orientation change along the closed path that does not alter the estimate, 
hence: $\min \left\{ \left| \theta_j - \theta_i - \pi/2 -k \pi \right| , \, k \in \Z \right\} $.
This robustness measure quantifies the tolerable orientation uncertainty and thus provides an objective measure to trade off localization precision versus noise robustness in singular point identification.

\section{Relation between robustness and path choice}
\begin{figure}
\begin{minipage}[b]{0.05 \textwidth}
\textbf{a}
\vspace*{38mm} \\
\textbf{b} \vspace*{2mm}
\end{minipage}
\hspace{-0.05 \textwidth}
\includegraphics[trim= 5mm 0mm 15mm 5mm, clip=true, width=0.45 \textwidth, keepaspectratio]{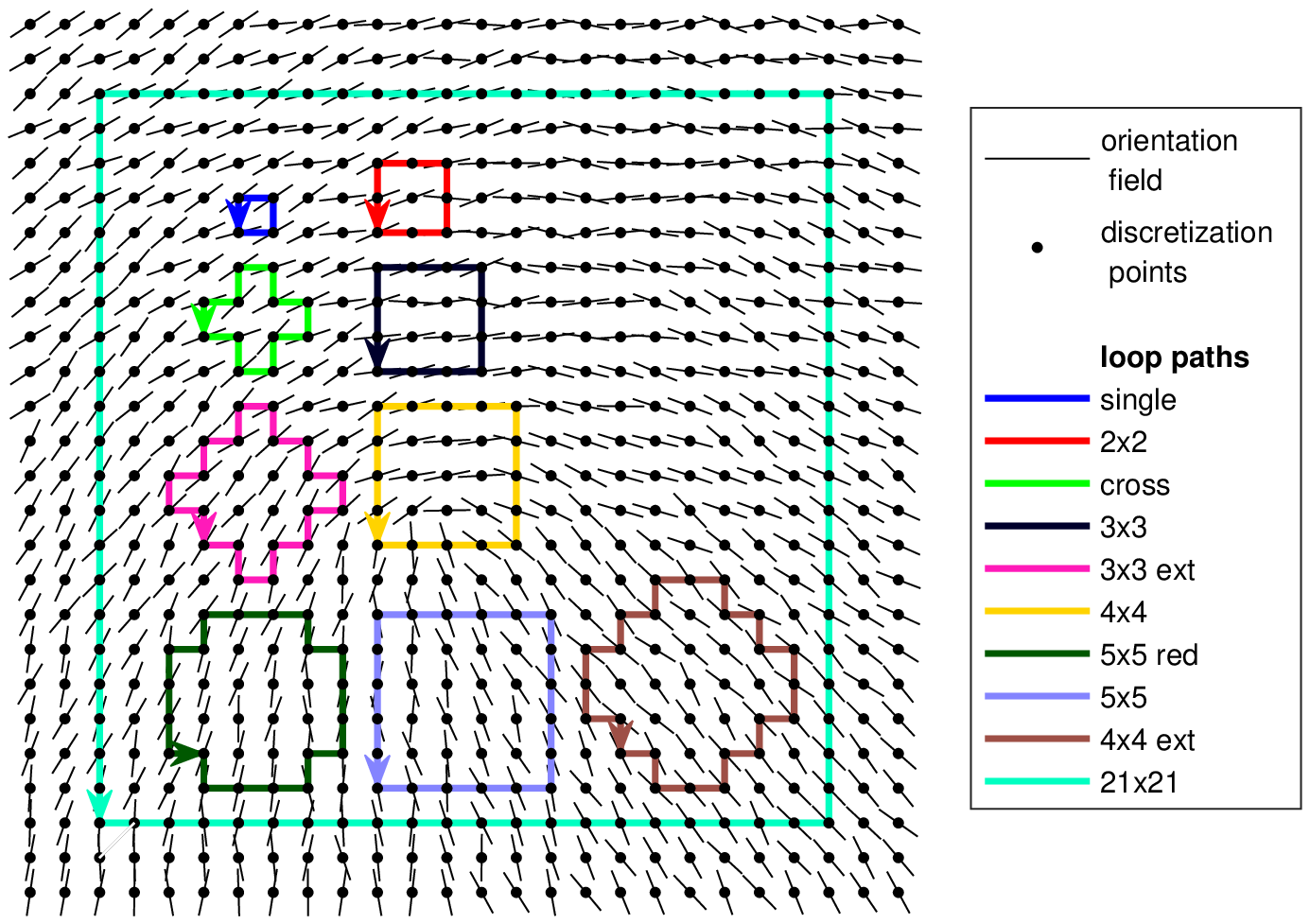}
\begin{minipage}[b]{0.05 \textwidth}
\textbf{c} 
\vspace*{38mm} \\
\textbf{d} \vspace*{2mm}
\end{minipage} 
\hspace{0.003 \textwidth}
\includegraphics[trim= 0mm -4mm 5mm 10mm, clip=true, width=0.448 \textwidth, keepaspectratio]{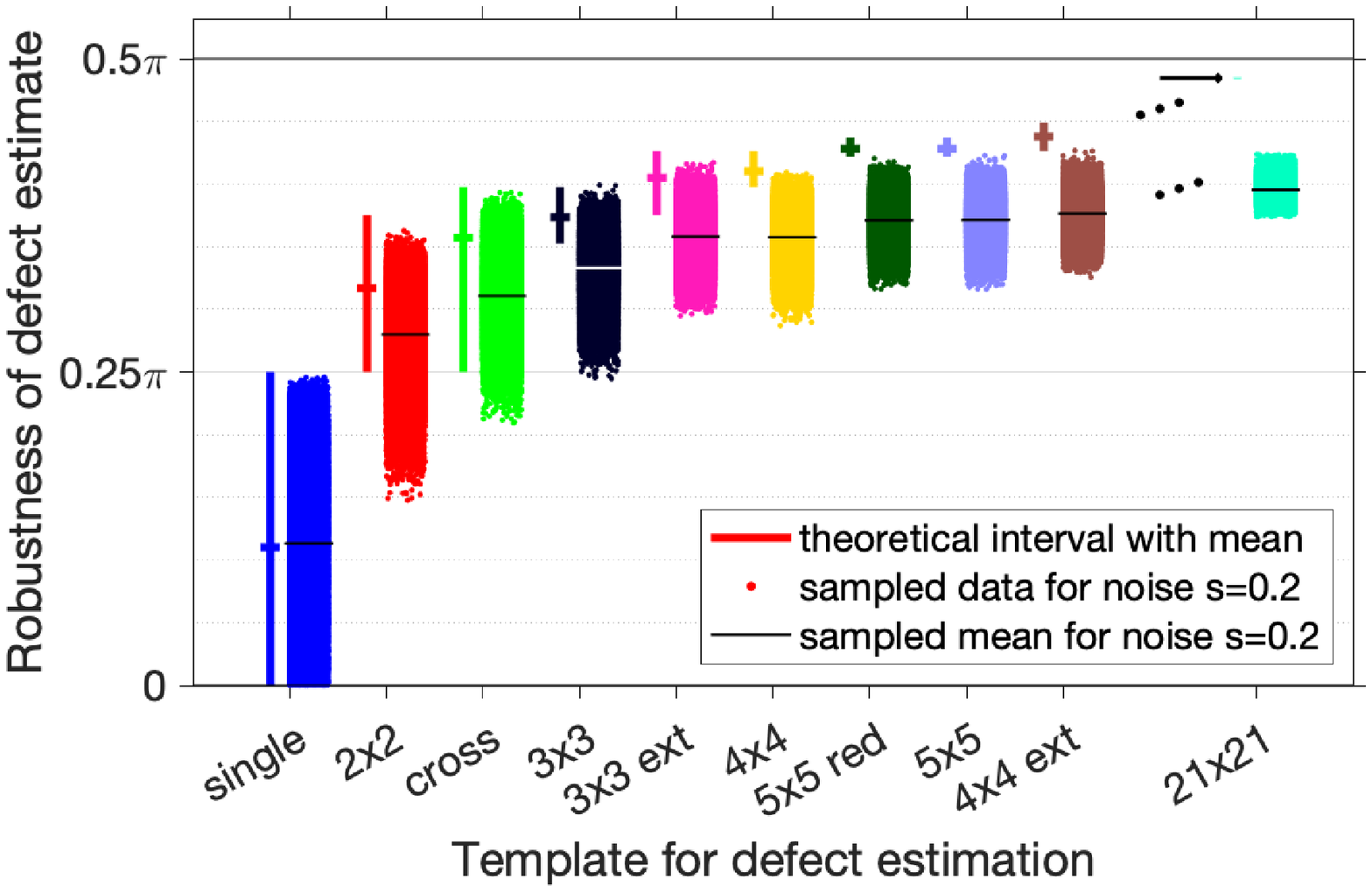} \\
\includegraphics[trim= 0mm 0mm 5mm 10mm, clip=true, width=0.47 \textwidth, keepaspectratio]{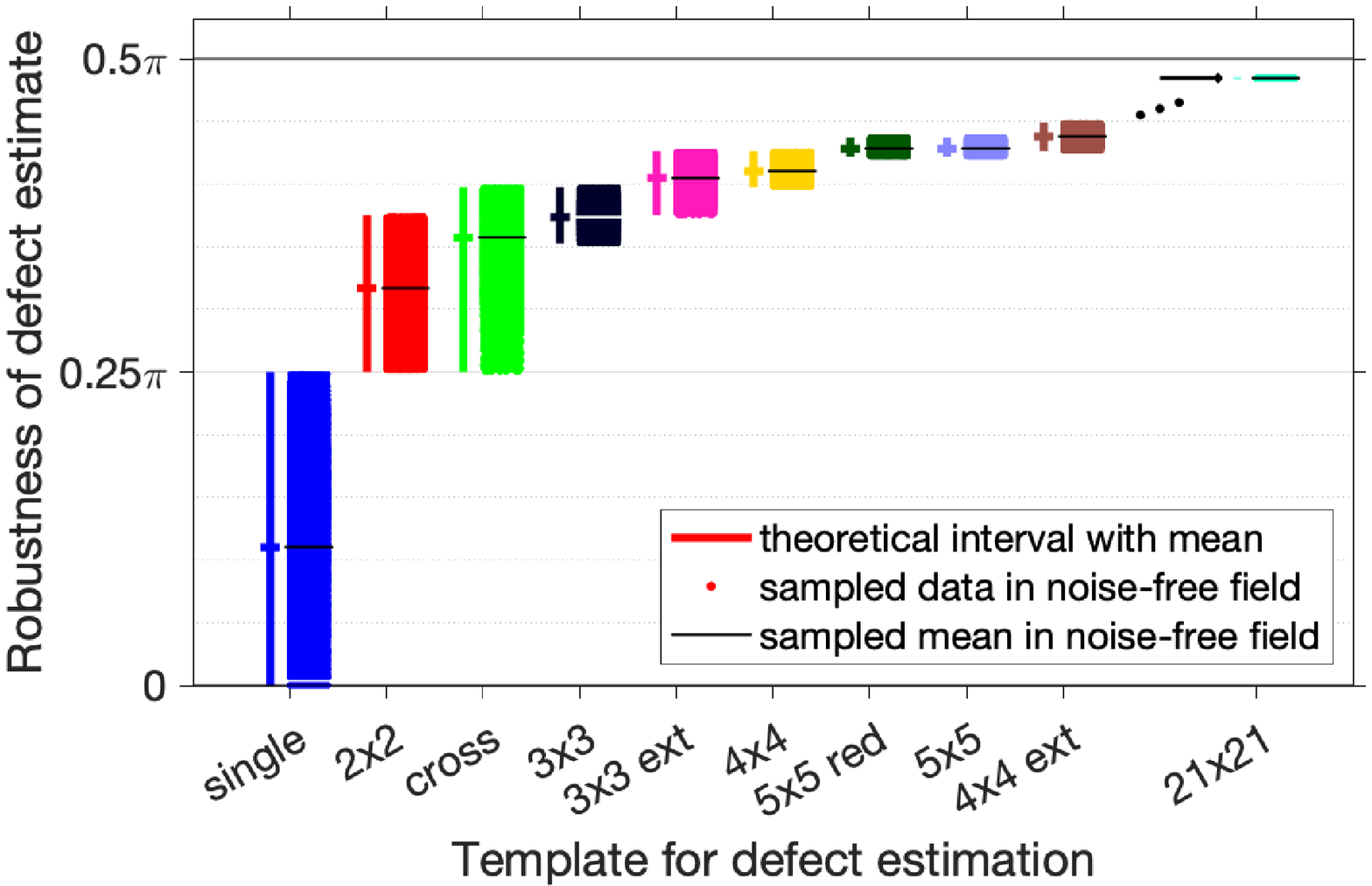}
\hspace{0.03 \textwidth}
\includegraphics[trim= 0mm 0mm 5mm 10mm, clip=true, width=0.47 \textwidth, keepaspectratio]{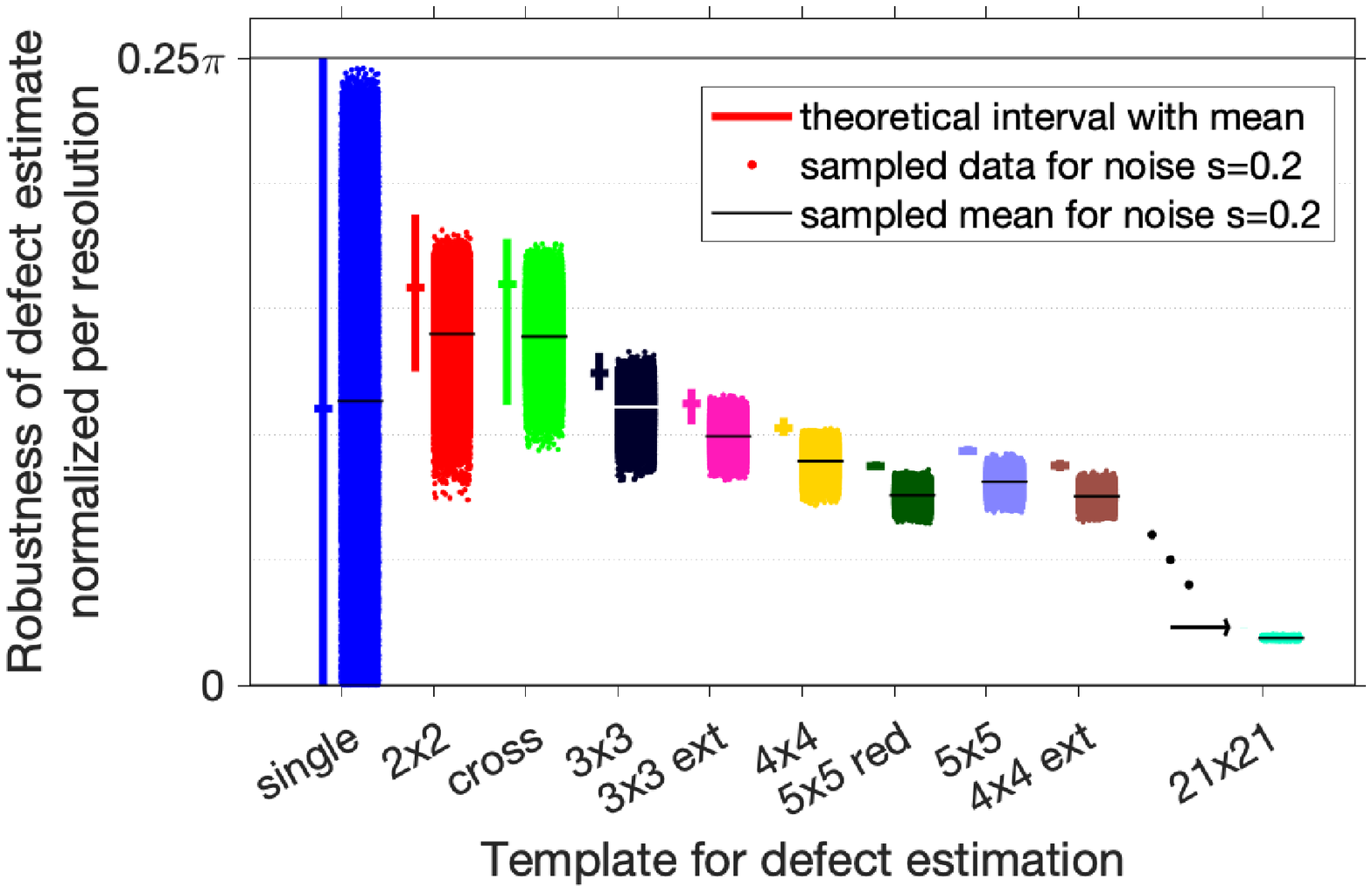}

\caption{Robustness of topological defect estimation; scatter plots: sampled data, solid lines: theoretical intervals of exact robustnesses. 
Arrows highlight smallest theoretical intervals.
\textbf{a} Color-coded template paths for defect estimation in a discretized orientation field with one $+1/2$ defect.
\textbf{b} Robustnesses in a noise-free field for 10,000 different locations of the true defect center. 
\textbf{c} Theoretical and empirical robustnesses for 10 independent realizations of uniform orientation noise with amplitude $s=0.2\,\text{rad}$ on each individual vector.
\textbf{d} Robustnesses with $s=0.2\,\text{rad}$ noise, normalized by the spatial resolution~$\approx \sqrt{\# \textnormal{pixels} }$ enclosed by the path. 
}
\label{fig:templatesAndRobustness}
\end{figure}

We use the above robustness measure to study how the choice of closed path used in defect estimation influences the estimator robustness, ultimately quantifying the trade-off between robustness and localization precision.
We consider a prototypical $+1/2$ defect and compare different template paths for defect estimation (Fig.~\ref{fig:templatesAndRobustness}\textbf{a}). To study robustness, we  
add noise in angle space as $ \theta_i \mapsto \theta_i + \delta \theta_i$ with $\delta \theta_i$ uniformly distributed in the interval~$\left[-s, +s\right]$. 

For noise-free orientation fields, the estimated robustnesses for the different path templates match the exact, theoretical values (Fig.~\ref{fig:templatesAndRobustness}\textbf{b}).
Both the upper and lower limits of robustness converge from below to the maximum possible value of $\pi/2$ when templates cover a circle of increasing diameter~$R$, which we prove as follows: 
An edge between neighboring discretization points has length~$h$ and is therefore seen from the defect center at a view angle $\alpha \leq \arcsin \left( \frac{h}{R} \right) $.
Then increasing template sizes ($R \to \infty$) implies $\alpha \to 0$ and hence a robustness of $\pi/2$. 
This limit agrees with vanishing distortions of vector fields by topological defects at large distance.

For noisy orientation fields, we observe that the estimator robustness decreases and its variance increases compared to the noise-free case. 
Even for a relatively large noise amplitude of $s=0.2\,\text{rad}$, however, the estimated robustnesses overlap with the exact, theoretical intervals for short paths (Fig.~\ref{fig:templatesAndRobustness}\textbf{c}). 
For long paths, robustnesses are systematically underestimated. 

The spatial resolution of defect localization scales with the diameter of the template, which is approximately the square root of the number of pixels enclosed by the path.
Normalizing the robustnesses by this resolution measure, we find an optimal trade-off between robustness and resolution:
As the smallest robustness for the chosen path should be maximal, the ``cross'' template is the best choice for $s=0.2\,\text{rad}$ noise (Fig.~\ref{fig:templatesAndRobustness}\textbf{d}). 
For different noise amplitudes (data not shown),  it is either the ``2x2'', ``cross'', ``3x3'', or ``3x3 ext'' path that is optimal. 
In noise-free fields, the ``2x2'' path is optimal (Fig.~\ref{fig:templatesAndRobustness}\textbf{d}, solid lines). 

In previous works,``2x2'' paths were mostly used~\cite{BazenGerez2002,DeCampRednerBaskaranHaganDogic2015}.
We can only speculate whether this was serendipitous, as the ``2x2''~naturally arises from the use of three-point finite differences, or because authors experienced trouble with the  ``single'' path, or whether it resulted from robustness considerations similar to ours.
The robustness measure proposed here enables quantitative answers to such questions and leads towards uncertainty quantification of defects in discretized orientation fields.

\newpage

\end{document}